\PassOptionsToPackage{prologue,table, xcdraw}{xcolor}
\documentclass[sigconf,screen]{acmart}
\usepackage{array}
\usepackage{CJKutf8}
\usepackage[table,xcdraw]{xcolor}

\usepackage{xspace}

\newcommand{\our}{LLaSA\xspace}

\AtBeginDocument{%
  }

\setcopyright{acmlicensed}
\copyrightyear{2024}
\acmYear{2024}
\acmDOI{XXXXXXX.XXXXXXX}

\acmConference[KDDCup '24]{Make sure to enter the correct
  conference title from your rights confirmation emai}{ Aug 28, 2024}{Barcelona, Spain}

\acmISBN{978-1-4503-XXXX-X/18/06}

\begin{document}

\title{LLaSA: Large Language and E-Commerce Shopping Assistant}

\author{Shuo Zhang}
\affiliation{%
  \institution{Zhejiang University}
  \city{Hangzhou}
  \country{China}
}
\email{shuo.zhang@zju.edu.cn}

\author{Boci Peng}
\affiliation{%
 \institution{Peking University}
 \city{Beijing}
 \country{China}}
\email{bcpeng@stu.pku.edu.cn}

\author{Xinping Zhao}

\affiliation{%
  \institution{Harbin Institute of Technology (Shenzhen)}
  \city{Shenzhen}
  \country{China}
}
\email{zhaoxinping@stu.hit.edu.cn}

\author{Boren Hu}
\affiliation{%
  \institution{ The Hong Kong University of Science and Technology (Guangzhou)}
  \city{Guangzhou}
  \country{China}
}
\email{huboren99@gmail.com}

\author{Yun Zhu}
\affiliation{%
  \institution{Zhejiang University}
  \city{Hangzhou}
  \country{China}
}
\email{zhuyun_dcd@zju.edu.cn}

\author{Yanjia Zeng}
\affiliation{%
  \institution{Zhejiang University}
  \city{Hangzhou}
  \country{China}
}
\email{22151306@zju.edu.cn}

\author{Xuming Hu}
\authornote{Xuming Hu is the corresponding author.}
\affiliation{%
  \institution{The Hong Kong University of Science and Technology (Guangzhou)}
  \city{Guangzhou}
  \country{China}
}
\email{xuminghu@hkust-gz.edu.cn}

\renewcommand{\shortauthors}{Zhang et al.}

\begin{abstract}
The e-commerce platform has evolved rapidly due to its widespread popularity and convenience. Developing an e-commerce shopping assistant for customers is crucial to aiding them in quickly finding desired products and recommending precisely what they need. 
However, most previous shopping assistants face two main problems: (1) \emph{task-specificity}, which necessitates the development of different models for various tasks, thereby increasing development costs and limiting effectiveness; 
and (2) \emph{poor generalization}, where the trained model performs inadequately on up-to-date products.
To resolve these issues, we employ Large Language Models (LLMs) to construct an omnipotent assistant, leveraging their adeptness at handling multiple tasks and their superior generalization capability. Nonetheless, LLMs lack inherent knowledge of e-commerce concepts.
To address this, we create an instruction dataset comprising 65,000 samples and diverse tasks, termed as \textbf{\textsc{EshopInstruct}}\footnote{Our instruction dataset can be found at \url{https://github.com/suyan-liang/EshopInstruct}.}. 
Through instruction tuning on our dataset, the assistant, named \textbf{\our}, demonstrates the potential to function as an omnipotent assistant. Additionally, we propose various inference optimization strategies to enhance performance with limited inference resources. 
In the Amazon KDD Cup 2024 Challenge\footnote{\url{https://www.aicrowd.com/challenges/amazon-kdd-cup-2024-multi-task-online-shopping-challenge-for-llms}}, our proposed method, \our, achieved an overall ranking of 3rd place on ShopBench, including 57 tasks and approximately 20,000 questions, and we secured top-5 rankings in each track, especially in track4, where we achieved the best performance result among all student teams. 
Our extensive practices fully demonstrate that LLMs possess the great potential to be competent e-commerce shopping assistants\footnote{All of the authors contributed equally to this work. Our team name is ``shimmering\_as\_the\_stars'', whose expression in Chinese is \begin{CJK}{UTF8}{gbsn}{灿若星辰}\end{CJK}.}.
\end{abstract}




\maketitle

\section{INTRODUCTION}
\subsection{Background}
The rapid growth of e-commerce has transformed how we shop, offering unprecedented convenience and access to a vast array of products. However, this convenience comes with the challenge of navigating an overwhelming volume of information. 
When shopping online, users often face the daunting task of sifting through countless products, reading numerous reviews, comparing prices, and ultimately making a purchase decision. This process can be time-consuming and stressful, highlighting the complexities inherent in online shopping~\cite{ref:ecomgpt,ref:ecomgpt-ct,ecinstruct}.
Large language models (LLMs) offer a promising solution to address these challenges~\cite{ref:llm}. Current techniques often struggle to fully grasp the nuances of specific shopping terms, customer behaviors, and the diverse nature of products and languages. 
In contrast, LLMs, with their multi-task and few-shot learning capabilities, have the potential to enhance the online shopping experience significantly. 
To encourage LLMs to meet the unique needs of online shopping, enhance user experience, and streamline decision-making, Amazon has introduced ShopBench and organized the Amazon KDD Cup 2024 challenge.
This competition features five tracks, focusing on four key shopping skills: Shopping Concept Understanding, Shopping Knowledge Reasoning, User Behavior Alignment, and Multilingual Abilities.
\begin{table*}
\renewcommand\arraystretch{1.2}
  \caption{Statistics of Shopbench.}
  \label{shopbench_sta}
  \begin{tabular}{lccccccc}
    \toprule
        \textbf{Track} & \textbf{\# Tasks} & \textbf{\# Questions} & \textbf{\# Products} & \textbf{\# Product Category} & \textbf{\# Attributes} & \textbf{\# Reviews} & \textbf{\# Queries}  \\
            \midrule
        All & 57 & 20598 & $\sim$13300 & 400 & 1032 & $\sim$11200 & $\sim$4500 \\
        Track1 & 27 & 11129 & $\sim$1500 & 400 & 1032 & $\sim$9600 & 361 \\
        Track2 & 8 & 3117 & $\sim$1000 & 400 & $\sim$10 & / & 552 \\
        Track3 & 15 & 3973 & $\sim$4800 & / & / & 1600 & $\sim$3600 \\
        Track4 & 7 & 2379 & $\sim$6000 & / & / & / & $\sim$520 \\
            \bottomrule
  \end{tabular}
\end{table*}
\begin{figure}[t]
    \includegraphics[width=1\linewidth]{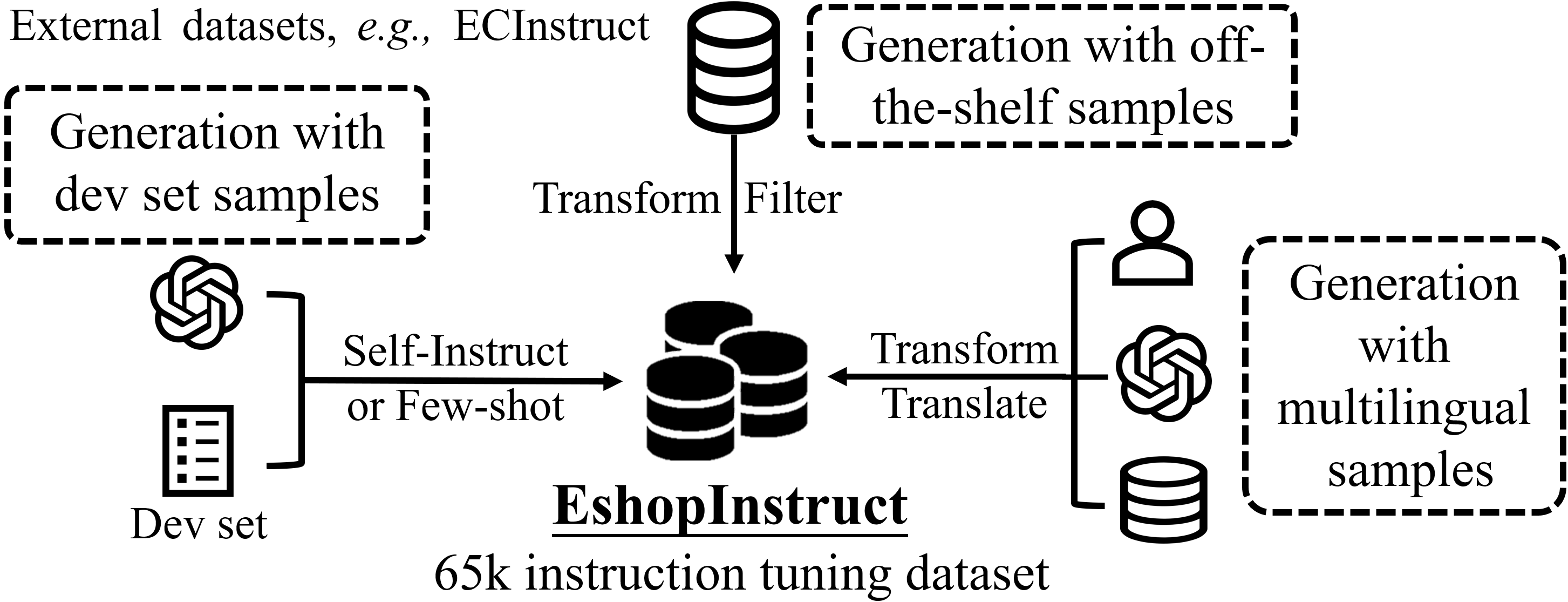} 
        \caption{Construction pipeline of \textsc{EshopInstruct}. We design three strategies for building the \textsc{EshopInstruct} dataset: generating data from seed data, extracting data from publicly available ECInstruct, and designing new tasks to generate data. Based on these strategies, we obtained 65k data points.} \label{data_construct}
\end{figure}

\subsection{Datasets Description}
ShopBench is a multi-task dataset derived from real-world shopping data in the Amazon platform, designed for the Amazon KDD Cup 2024 challenge. The dataset is divided into a few-shot development set and a test set, designed to more accurately simulate the few-shot learning settings. 
It contains 57 tasks and approximately 20,000 questions, which are all reformulated into a unified text-to-text generation format to facilitate LLM-based solutions. The detailed statistics of the datasets are summarized in Table~\ref{shopbench_sta}.

\subsection{Task Description}
In the ShopBench benchmark, five abilities, including Generation, Ranking, Retrieval, Multiple-Choice, and NER (Named Entity Recognition), are introduced to evaluate four important shopping skills:
\begin{itemize}
  \item Track1 (Shopping Concept Understanding): Given the prevalence of domain-specific concepts in online shopping, the goal is to enhance LLMs' ability to effectively understand and respond to queries about these concepts.
  \item Track2 (Shopping Knowledge Reasoning): Considering the complex reasoning required for shopping decisions, the goal is to assess the model's capability in reasoning about products and their attributes using domain-specific implicit knowledge.
  \item Track3 (User Behavior Alignment): Given the diversity and implicit nature of user behaviors in online shopping, the goal is to align language models with these behaviors to improve their effectiveness in this domain.
  \item Track4 (Multi-Lingual Abilities): Recognizing the need for multi-lingual models in online shopping, the goal is to evaluate a single model's performance across different shopping locales without re-training, focusing on multi-lingual concept understanding and user behavior alignment.
\end{itemize}
\begin{figure}[t]
    \centering
    \includegraphics[width=0.75\linewidth]{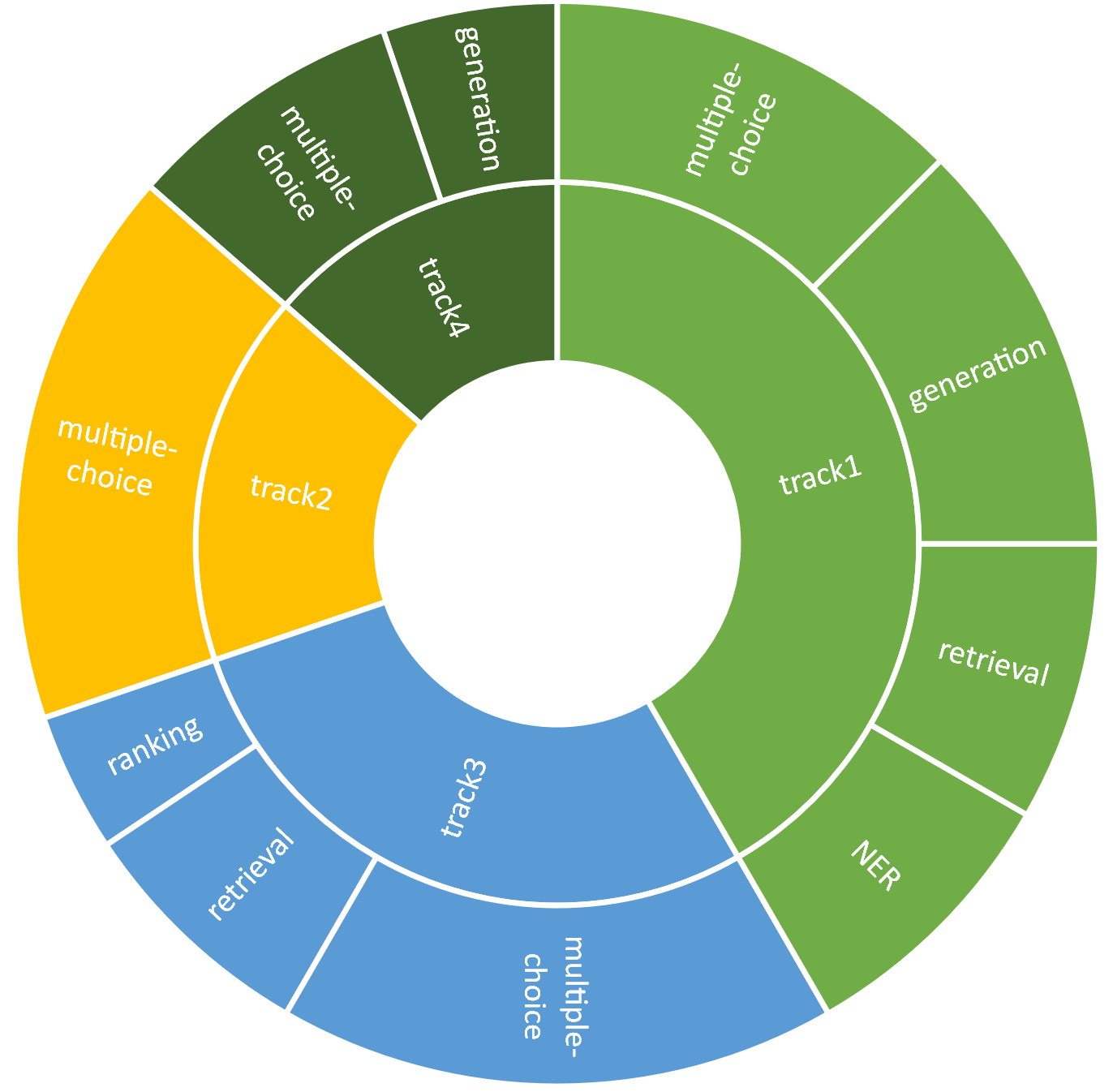} 
    \caption{The data distribution of development set, including four important shopping skills: (1) Shopping Concept Understanding; (2) Shopping Knowledge Reasoning; (3) User Behavior Alignment; (4) Multi-Lingual Abilities, and five abilities: (\lowercase\expandafter{\romannumeral1}) Generation; (\lowercase\expandafter{\romannumeral2}) Ranking; (\lowercase\expandafter{\romannumeral3}) Retrieval; (\lowercase\expandafter{\romannumeral4}) Multiple-Choice; (\lowercase\expandafter{\romannumeral5}) NER.} \label{shopbench}
\end{figure}

\section{TRAINING DATASET CONSTRUCTION}
While LLMs exhibit strong generalization across multiple tasks, they often perform poorly in specific domains due to a lack of relevant knowledge. This competition involves many tasks related to online shopping, and general-purpose models lack knowledge in this area. 
Therefore, directly adapting a general-purpose model to the online shopping scenario is quite challenging. To improve the model’s performance in this domain, we need to inject relevant knowledge into it.
In this challenge, the organizers did not provide a large-scale training dataset. As a result, we constructed our training dataset using publicly available data, our data construction pipeline is shown in Figure~\ref{data_construct}.

\subsection{Development Set Analysis}

We analyzed the provided development data to gain insights for constructing the training dataset. The development set comprises 96 data points across 18 different tasks, the distribution of task types is shown in Figure \ref{shopbench}.

\subsubsection{Shopping Concept Understanding}
This track focuses on evaluating the model's ability to understand entities and concepts specific to the online shopping domain, which can be divided into the following sub-tasks:
\begin{itemize}
    \item Concept Normalization: Given a product name, select the product that represents the same concept as the current product name.
    \item Elaboration: Given a concept, explain it in plain, understandable, and concise language.
    \item Extraction and Summarization: Extract and summarize the product names mentioned within the product description.
    \item Relation Inference: Given four options, select the product category that has a certain attribute.
    \item Concept Explanation: Describe the concept of the corresponding product.
    \item Sentiment Analysis: Select 3 snippets from a list that customers are most likely to write in their reviews.
\end{itemize}

\subsubsection{Shopping Knowledge Reasoning}
This track aims to evaluate the model's ability to understand complex implicit knowledge in the online shopping domain and to apply this knowledge to various types of reasoning:
\begin{itemize}
    \item Numerical Reasoning: Extract related numeric information and perform numeric reasoning.
    \item Commonsense Reasoning: Recommend daily products that are most likely to be purchased based on the current product in the shopping list.
    \item Implicit, Multi-Hop Reasoning: Understand the implicit, domain-specific knowledge and infer multi-hop relations between shopping entities.
\end{itemize}

\subsubsection{User Behavior Alignment}
This track aims to assess a model's ability to understand implicit relationships in user behavior, thereby enabling its recommendation capabilities.
\begin{itemize}
    \item Recommendation based on user queries: Given a list of product IDs, rank the products according to their relevance to the query.
    \item Behavior predictions: Given the user's previous actions, infer the next action.
    \item Recommendation based on purchase histories: Given the products a user has just purchased, predict what they might buy next.
    \item Sentiment Label Predictions: Given a comment, score it based on its sentiment.
\end{itemize}

\subsubsection{Multi-Lingual Abilities}
This track focuses on how models can extend their capabilities to multiple languages to simultaneously meet the needs of global markets.
\begin{itemize}
    \item Multilingual shopping concept understanding: The tasks in Track1 are expanded to multiple languages.
    \item Multilingual user behavior alignment: The tasks in Track3 are expanded to multiple languages.
\end{itemize}

\subsection{External Datasets}
We collect several external datasets related to Amazon products. Leveraging these datasets, we can construct various tasks and corresponding data tailored for SFT. These external datasets are listed as follows:
\begin{itemize}
    \item ECInstruct~\cite{ecinstruct}: It is an open-source SFT dataset in the e-commerce domain, encompassing 10 different tasks and comprising 264K instances, including Sequential Recommendation, Query Product Rank, etc.
    \item Amazon-ESCI~\cite{ref:esci}: It is a large-scale multilingual query-product dataset, which was employed in the KDD Cup 2022 competition. It includes three sub-tasks: Query-Product Ranking, Multi-class Product Classification, and Product Substitute Identification. 
    \item Amazon-M2~\cite{ref:amazon-m2}: It is a multilingual session-based recommendation dataset designed for the KDD Cup 2023 competition.
    \item Amazon Reviews 2023~\cite{ref:review}: It is a comprehensive Amazon product dataset that not only includes user reviews for various products but also provides extensive information such as brand, description, category, features, and co-purchase relationships. It is an updated version of Amazon Reviews 2018.
    \item OA-Mine \& AE-110K~\cite{ref:oa}: They are two NER datasets in the E-commerce domain, designed to extract categories, brands, target audiences, and other product characteristics from product names.
    \item Amazon-Category\footnote{\url{https://huggingface.co/datasets/ikeno-ada/amazon\_category}}: It provides various products along with their corresponding categories, encompassing items from multiple languages.
\end{itemize}
\subsection{Training Data Construction Strategy}
\begin{figure*}[ht]
    \includegraphics[width=1.0\linewidth]{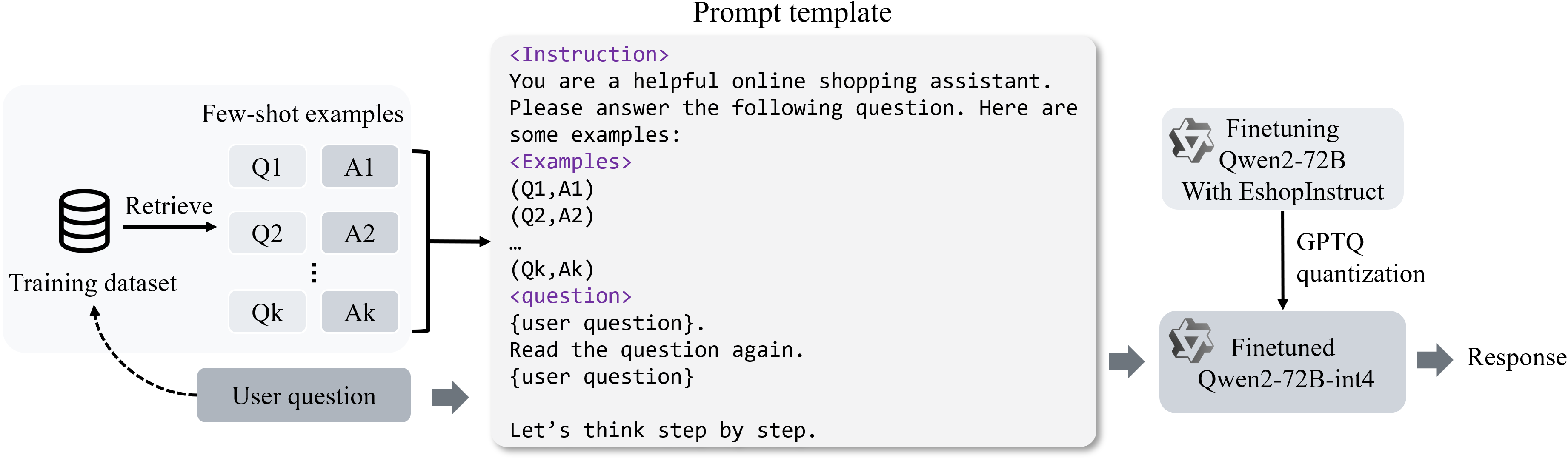} 
    \caption{The overall inference framework of our solution. 
    In our prompt construction, we enhanced the model's reasoning capabilities by incorporating few-shot examples retrieved using queries, the ``read again'' technique, and chain-of-thought reasoning. Additionally, for Qwen2-72B, we applied GPTQ quantization, enabling it to run efficiently on limited resources.} \label{inference}
\end{figure*}
We utilized public datasets and OpenAI's ChatGPT and GPT4\footnote{\url{https://openai.com/api/}} \cite{DBLP:journals/corr/abs-2303-08774} for the data construction of our \textsc{EshopInstruct}, whose detailed construction pipeline is shown in Figure \ref{data_construct}. 
Our data construction strategy can be categorized into three main approaches:
\begin{itemize}
    \item We analyzed 18 tasks and their corresponding data in the development set, using this analysis to generate more data that aligns more closely with the types found in the development set.
    \item Based on the practical scenarios of each track, we developed additional task types and corresponding data beyond those in the development set.
    \item To increase the proportion of the real-world data in the SFT dataset and provide more knowledge to the LLMs, we also created a substantial amount of data based on external datasets.
\end{itemize}
To create data that aligns with the task types in the development set, we adopted two strategies. Firstly, for task types that can be directly constructed or transformed using existing datasets like ECInstruct, we generated the corresponding data directly from these datasets. 
For example, tasks such as Elaboration, Extraction and Summarization, Relation Inference, Sentiment Analysis in Track1, and Recommendation based on query in Track3, we identified similar data in ECInstruct~\cite{ecinstruct}. For these tasks, we directly extracted data from ECInstruct and transformed them into the standard format. 
Secondly, for task types where it was challenging to extract data from existing datasets, we utilized LLMs, such as GPT-4, for data generation. 
For numeric reasoning, implicit and multi-hop reasoning in Track2, as well as user behavior prediction in Track3, we used GPT-4 for data construction. 
When using GPT-4 to generate this portion of the data, we provided the model with few-shot examples and employed the chain-of-thought method, enabling it to generate the reasoning process to ensure data quality.
Considering that the task types in the development set do not comprehensively cover all scenarios, we constructed additional tasks and corresponding data based on descriptions from various tracks. 
For instance, we observed that there is no relevant data about the Concept Normalization task in Track1 and the Daily Product Recommendation in Track2 in the development set. 
Therefore, we constructed corresponding data for them. These data constructions may involve transformations from external datasets or generation by LLMs. Additionally, we referred to the methods in Self-Instruct \cite{self-instruct} to generate a portion of the data. 
Specifically, we used development data as seed data and then utilized GPT-3.5-turbo to generate instructions and corresponding responses based on this data. Subsequently, we employed GPT-4 as a judge to filter the data.
Furthermore, given that much of the data constructed through the first two methods is generated by LLMs, there may be a considerable amount of noise, and scalability could be limited due to cost constraints. 
Therefore, to introduce a substantial amount of real-world data to our models, we have also constructed additional tasks and corresponding data from external datasets. For example, leveraging the Amazon-ESCI dataset, we constructed tasks such as Query Generation, Related Product Retrieval, etc. 
It should be noted that in order to enhance our model's multilingual processing capabilities, we incorporated a significant amount of data related to products in various languages other than English during the dataset construction phase.
Following the above strategy, we ultimately obtained approximately 65,000 data entries in \textsc{EshopInstruct}. 
Moreover, to further augment our training dataset, we strategically sampled a subset of data from the ECInstruct dataset. We then used these data for instruction tuning.
\section{INSTRUCTION TUNING}
We use instruction tuning to incorporate online shopping-related knowledge into the LLMs and enhance their instruction-following capabilities. 
Given the size of our constructed dataset (65,000 entries) and our limited training resources, we adopted the LoRA (Low-Rank Adaptation)~\cite{lora} fine-tuning method, following the standard approach of auto-regressive language modeling.
During phases 1 and 2 of the challenge, we experimented with four models\footnote{All models used are chat or instruct models} of different sizes: Mistral-7B\footnote{\url{https://huggingface.co/mistralai/Mistral-7B-Instruct-v0.2}}~\cite{jiang2023mistral}, LLama3-8B\footnote{\url{https://github.com/meta-llama/llama3}}~\cite{ref:llama3}, and Qwen2-7B/72B\footnote{\url{https://github.com/QwenLM/Qwen2}}~\cite{qwen2}. 
Some key training hyper-parameters are listed in Table~\ref{hyper}. 
We used the standard AdamW optimizer \cite{DBLP:conf/iclr/LoshchilovH19} for supervised fine-tuning (SFT) optimization, with a cosine learning rate schedule, a peak learning rate of $4\times \mathrm{e}^{-5}$, and a 10\% warmup ratio. All the models were trained with multiple NVIDIA A800 80G GPUs.
For models with fewer than 10 billion parameters, such as Mistral-7B, LLama3-8B, and Qwen2-7B, we trained on a single GPU without quantization.
For the Qwen2-72B model, we used bf16 precision for LoRA fine-tuning, employed DeepSpeed's ZeRO Stage3 \cite{DBLP:conf/sc/RajbhandariRRH20} for fine-tuning across four GPUs, and then used GPTQ to quantize the parameters to 4-bit precision.
\begin{table}
    \renewcommand\arraystretch{1.2}
    \centering
    \caption{Hyperparameter for Instruction-tuning.}
    \begin{tabular}{cc}
    \toprule
    \textbf{Configuration} & \textbf{Value} \\ 
    \midrule
        Model & Qwen2-72B \\
        Number of epochs & 2\\
        Learning Rate & $4e$-$5$ \\
        Max Length & 2048 \\
        Devices & 8 NVIDIA A800 GPUs (80GB) \\
        LR Scheduler & Cosine \\
        Warmup Raion & 0.1 \\
        Total Batch Size & 256 \\
        Optimizer & AdamW \cite{DBLP:conf/iclr/LoshchilovH19} \\
        Lora Rank & 8 \\
        Lora Target & $\mathrm{q}_{proj}, \mathrm{k}_{proj}, \mathrm{v}_{proj}$\\
    \bottomrule
    \label{hyper}
    \end{tabular}
\end{table}

\section{INFERENCE}
In this section, we will introduce our inference strategy, which comprises two key components: quantization and prompt engineering. Our overall inference strategy pipeline is shown in Figure~\ref{inference}.
\subsection{Quantification}
During the challenge, the submission will run on a T4 GPU with 16GB of memory. In Phase 2, four T4 GPUs will be provided, which means only 64GB of GPU memory will be available.
To experiment with larger models (such as models with 72B parameters) and minimize the reduction in model capability while reducing the required GPU memory as much as possible, we leveraged quantization techniques. 
Specifically, we adopted GPTQ quantization \cite{frantar-gptq}, a post-training quantization method where each row of the weight matrix is independently quantized to int4 to reduce error but restored to fp16 during inference for better performance.
In this challenge, we only applied quantization to the Qwen2-72B model. After training, we utilized 1,000 data samples from Alpaca \cite{alpaca} generated by GPT-4 for quantization calibration of the model.

\subsection{Prompting Strategies}
Through the analysis of the development set, we found that many tasks in the challenge involve reasoning. To improve the model's performance on these tasks, we introduced Chain of Thought (CoT) \cite{cot}, a technique that can significantly enhance the complex reasoning abilities of large language models. we implemented a simple zero-shot Chain-of-Thought in our solution. 
Additionally, we retrieved the three most relevant samples from the constructed training dataset as few-shot examples, which yielded better results.
The test data can be roughly divided into multiple-choice and non-multiple-choice types. We adopt different processing measures and prompts for these two types of data. 
For questions that may involve reasoning, we encourage the model to think more deeply and use regular expressions to extract the final answer. 
For generation-related questions, we let the model directly output the final result.
Considering the importance of user input in online shopping scenarios, we have also implemented a simple and effective prompting method called Re-Reading \cite{xu2023re} which entails re-reading the question to enhance reasoning capabilities in Large Language Models.
\begin{table*}
\renewcommand\arraystretch{1.2}
  \caption{Overall performance on different tracks, where the best results are {boldfaced} and the second-best results are {underlined}. ``-'' denotes missing experimental data due to the unstable evaluation system.}
  \label{results}
  \begin{tabular}{lcccccc}
    \toprule
        \textbf{Model}                & \textbf{Quantization} & \textbf{Track1} & \textbf{Track2} & \textbf{Track3} & \textbf{Track4} & \textbf{Track5}  \\
            \midrule
         \textbf{(A)} Mistral-7B+ECInstruct          & none              & 0.702           & 0.529           & 0.602           & 0.635           & -               \\
         \textbf{(B)} LLama3-8B+\textsc{EshopInstruct}              & none              & 0.741           & 0.6             & 0.625           & 0.651           & 0.67            \\
         \textbf{(C)} Qwen2-7B+\textsc{EshopInstruct}            & none              & 0.744           & 0.629           & 0.614           & 0.664           & -               \\
        \midrule \midrule
         \textbf{(D)} Qwen2-72B           & GPTQ-int4            & 0.786           & 0.716           & \underline{0.706}           & 0.654           & 0.722           \\
         \textbf{(E)} Qwen2-72B+ECInstruct & GPTQ-int4             & \underline{0.801} & \underline{0.719}  & 0.703  &  \underline{0.686}  & \underline{0.747}  \\
         \textbf{(F)} Qwen2-72B+\textsc{EshopInstruct} & GPTQ-int4            & \textbf{0.824}  & \textbf{0.747}  & \textbf{0.713}  & \textbf{0.735}  & \textbf{0.763}  \\
    \bottomrule
  \end{tabular}
\end{table*}
\begin{table*}[ht]
    \renewcommand\arraystretch{1.2}
    \centering
    \caption{Best Model's Detailed Performance on different task types. ``-'' means that this track does not evaluate this type of task.}
    \begin{tabular}{lcccccc} 
    \toprule
    \textbf{Track} & \textbf{Generation} & \textbf{Multiple-Choice} & \textbf{NER} & \textbf{Retrieval} & \textbf{Ranking} & \textbf{Overall} \\ 
    \midrule
        Track1 & 0.732 & 0.860~ & 0.789 & 0.858 & - & 0.824 \\
        Track2 & - & 0.770~ & - & 0.588 & - & 0.747 \\
        Track3 & 0.618 & 0.695 & - & 0.813 & 0.845 & 0.713 \\
        Track4 & 0.482 & 0.838 & - & - & 0.83 & 0.735 \\
        Track5 & 0.763 & 0.793 & 0.802 & 0.765 & 0.659 & 0.844 \\
    \bottomrule
    \label{detail}
    \end{tabular}
\end{table*}
\section{RESULTS}
In this section, we will compare and analyze the performance of different models across the five tracks, as shown in Table \ref{results}. Overall, the size of the model parameters has a considerable impact on performance, with larger parameter models generally performing better.
\textbf{(A)} shows relatively weaker performance across Track1 to Track4, especially on Track2, where it scored only 0.529. In comparison, \textbf{(B)} shows improved performance across all tracks, particularly on Track1 and Track5. 
Despite having the same 7B parameters as \textbf{(A)}, \textbf{(C)} performs well across all available tracks, especially on Track2 and Track4. Comparing \textbf{(B)} and \textbf{(C)}, we find that Qwen2-7B owns a greater potential than LLama3-8B in providing e-commerce shopping assistance. 
Therefore, we chose the Qwen2 series as our backbone model since it performs relatively well.
With 72B parameters, \textbf{(D)} demonstrates excellent performance across all tracks, particularly on Track1 to Track3, achieving high scores of 0.786, 0.716, and 0.706, respectively. However, its performance slightly drops on Track4 to 0.654, but it remains at a high level.
To incorporate the domain knowledge of e-commerce,  we fine-tuned Qwen2-72B on ECInstruct and our constructed \textsc{EshopInstruct}, respectively.
Comparing \textbf{(E)} and \textbf{(F)}, we can see that \textbf{(F)} consistently and considerably outperforms \textbf{(E)} in all tracks, indicating the superiority of supplementing e-commerce shopping tasks with our \textsc{EshopInstruct}. 
It is worth mentioning that we find LLMs fine-tuned on ECInstruct perform badly on generation tasks. The performance of \textbf{(F)} on specific tasks is detailed in Table \ref{detail}. 
By utilizing our carefully constructed training dataset \textsc{EshopInstruct} for instruction fine-tuning and employing effective inference strategies, we ultimately secured 3rd place overall in the Amazon KDD Cup 2024 Challenge, 3rd place in Track1, 2nd place in Track4, and ranked within the top 5 for the remaining tracks.
\section{CONCLUSION}
In this paper, we present our solution for the Amazon KDD Cup 2024 Challenge. We constructed a multi-task instruction dataset called \textbf{\textsc{EshopInstruct}}, which contained 65,000 samples tailored to online shopping scenarios. 
In addition, we utilized \textbf{\textsc{EshopInstruct}} for instruction tuning on large language models, resulting in knowledgeable shopping assistants named \textbf{{LLaSA}}. 
To optimize inference performance with limited resources, we employed GPTQ quantization and prompting strategies such as Chain-of-Thought and Re-Reading. 
Evaluation results demonstrated the effectiveness of our approach, securing 3rd place on the overall leaderboard and ranking within the top 5 in each track.
Especially in track4 (Multi-lingual Abilities), we obtained the best student team award. 
\bibliographystyle{ACM-Reference-Format}
\bibliography{sample-base}

\end{document}